\documentclass{article}
\usepackage{ijcai22}

\usepackage{times}
\usepackage{soul}
\usepackage{url}
\usepackage[hidelinks]{hyperref}
\usepackage[utf8]{inputenc}
\usepackage[small]{caption}
\usepackage{graphicx}
\usepackage{amsmath}
\usepackage{amsthm}
\usepackage{booktabs}
\usepackage[ruled]{algorithm2e}
\title{Federated Learning on Heterogeneous and Long-Tailed Data via Classifier Re-Training with Federated Features}

\author{
Xinyi Shang$^{\rm 1}$
\and 
Yang Lu$^{\rm 1}$\thanks{Corresponding author: Yang Lu, luyang@xmu.edu.cn}
\and
Gang Huang$^{\rm 2}$
\And
Hanzi Wang$^{\rm 1}$
\affiliations
$^{\rm 1}$Fujian Key Laboratory of Sensing and Computing for Smart Cities, School of Informatics, \\Xiamen University, Xiamen, China\\
$^{\rm 2}$ Zhejiang Lab, Hangzhou, China
\emails
shangxinyi@stu.xmu.edu.cn, luyang@xmu.edu.cn, huanggang@zju.edu.cn, hanzi.wang@xmu.edu.cn
}

\usepackage{bm}
\usepackage{amssymb}
\usepackage{booktabs}
\usepackage{multirow}
\usepackage[table,xcdraw]{xcolor}
\usepackage{amsmath}
\usepackage{xcolor}
\usepackage{subcaption}

\def\bx{\mathbf{x}}
\def\bs{\mathbf{s}}

\def\bw{\mathbf{w}}
\def\bz{\mathbf{z}}
\def\bv{\mathbf{v}}
\def\DD{\mathcal{D}} 
\def\CC{\mathcal{C}} 
\def\DS{\mathcal{S}} 

\def\bu{\mathbf{u}}
\def\bv{\mathbf{v}}
\def\bg{\mathbf{g}}
\def\IF{\textrm{IF}}
\begin{document}

\maketitle

\begin{abstract}
	Federated learning (FL) provides a privacy-preserving solution for distributed machine learning tasks. One challenging problem that severely damages the performance of FL models is the co-occurrence of data heterogeneity and long-tail distribution, which frequently appears in real FL applications. In this paper, we reveal an intriguing fact that the biased classifier is the primary factor leading to the poor performance of the global model. Motivated by the above finding, we propose a novel and privacy-preserving FL method for heterogeneous and long-tailed data via Classifier Re-training with Federated Features (CReFF). The classifier re-trained on federated features can produce comparable performance as the one re-trained on real data in a privacy-preserving manner without information leakage of local data or class distribution. Experiments on several benchmark datasets show that the proposed CReFF is an effective solution to obtain a promising FL model under heterogeneous and long-tailed data. Comparative results with the state-of-the-art FL methods also validate the superiority of CReFF. Our code is available at \color{magenta}\url{https://github.com/shangxinyi/CReFF-FL}\color{black}. 
\end{abstract}

\section{Introduction}
The emergence of federated learning (FL) enables multiple clients to collaboratively learn a powerful global model without transmitting local private data to the server.
%The clients send their models to the server, and then the global model is updated by aggregating the received models on the server.
Serving as a communication-efficient and privacy-preserving learning framework, FL has shown its potential to facilitate real-world applications in multiple domains, e.g., natural language processing \cite{jiang2021federated} and fraudulent credit card detection \cite{zheng2020federated}. 
%healthcare \cite{9488776}, 
However, during FL model training, one major practical challenge is data heterogeneity \cite{li2020federated}. In the FL environment, data generated or collected by different clients may come from various sources, which results in distribution discrepancy between clients.
Moreover, real-world data often exhibits long-tail distribution with heavy class imbalance, where the number of samples in some classes (called head classes) severely outnumbers that in some other classes (called tail classes).
Building an unbiased classification model on data with this kind of distribution is termed long-tail learning, which has been extensively studied in recent years \cite{2021deeplt}. 
In the FL scenario, if the training data across clients is long-tailed and heterogeneous at the same time, the joint problem becomes complicated and challenging because each client may hold different tail classes.
For example, medical institutions aim to build a diagnosis model on private patient records, which are held by each institution locally. Each institution has different disease distributions, and different diseases make up long-tail distribution because some diseases are common and others are rare. Adopting FL in this case, local models will poorly perform on different tail classes due to data heterogeneity and long-tail distribution, and the aggregated global model will also be affected.
One straightforward way is to simply adopt existing long-tail learning methods to the joint problem in FL. 
However, most of these methods require the information of local or global class distribution as prior knowledge for optimization, which may expose potential privacy issues.

%The existing solutions for non-IID data in federated learning generally perform poorly on the tail classes due to the lack of consideration of the universal long-tail distribution.
\begin{figure}[!t]
	\centering
	\begin{subfigure}[b]{0.48\linewidth}
		\includegraphics[width=\linewidth]{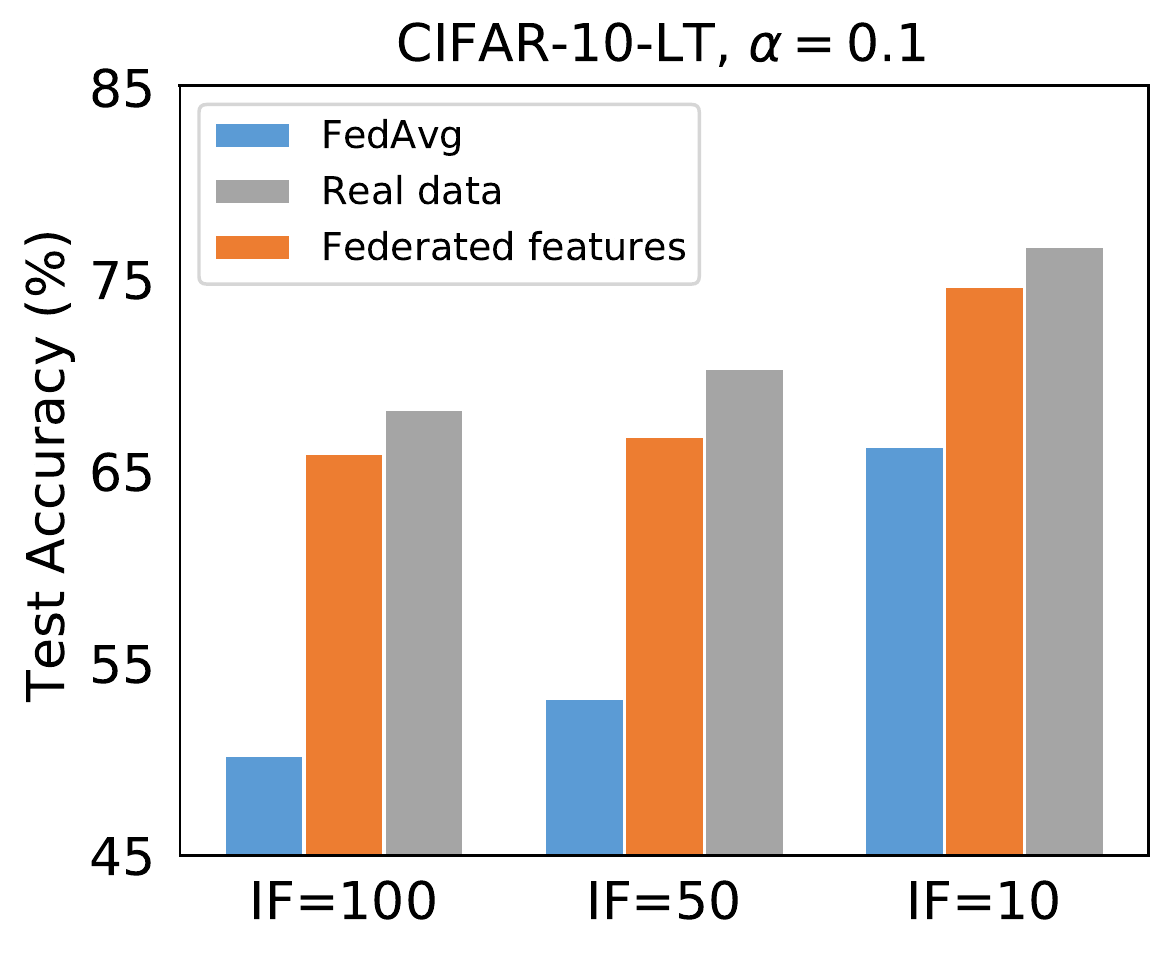}
	\end{subfigure}
	\begin{subfigure}[b]{0.48\linewidth}
		\includegraphics[width=\linewidth]{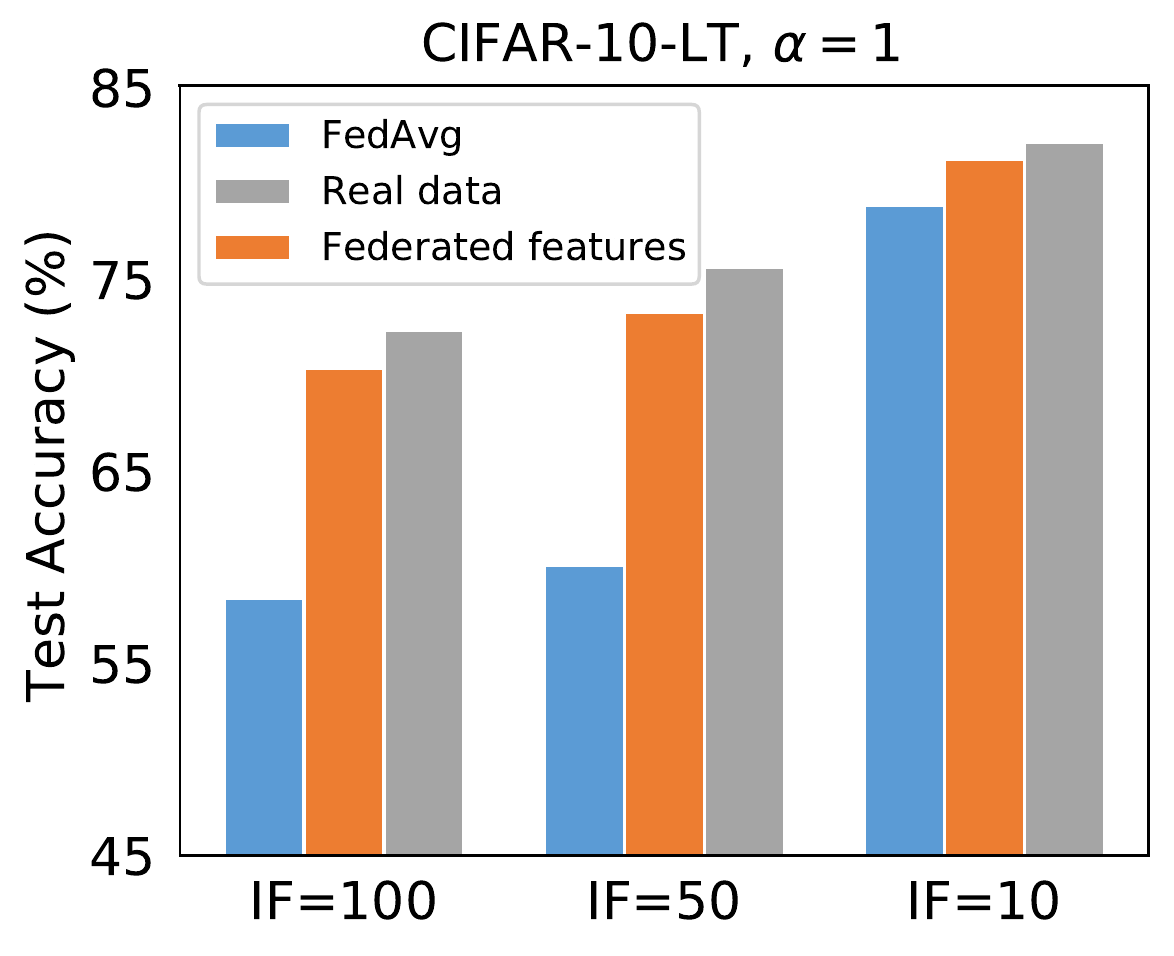}
	\end{subfigure}
	\caption{The performance of re-training the classifier of a FedAvg model using the same amount of data but from different sources. Real data is directly collected from clients, and federated features are synthesized by the proposed CReFF.}
	\label{motivation_figure}
	%\vspace{-10px}
\end{figure}
%%提出只用数据去训练分类器就能得到很好的效果
One of the recent advances in long-tail learning is decoupling the training process into representation learning and classifier re-training \cite{Kang2020Decoupling}. That is, with the trained feature extractor fixed, a biased classifier is re-trained using a set of balanced data, which is also called two-stage learning. In the FL framework, an FL model can also be decoupled into two-stage training, where the feature extractor is obtained by aggregating client models and the classifier is re-trained on the server using a set of balanced data.
We conduct a quick experiment on CIFAR-10-LT (the long-tailed version of CIFAR-10 \cite{NEURIPS2019_621461af}) to show that two-stage learning performs well on heterogeneous and long-tailed data in FL. We use imbalance factors ($\IF$) and the Dirichlet distribution coefficient $\alpha$ to control the degree of long-tail and heterogeneity, respectively. Higher $\IF$ means a higher degree of imbalance degree, and lower $\alpha$ means a higher degree of heterogeneity.
%In this work, we conduct a quick experiment to re-train the global model classifier on a shared balanced data collected from clients. Note that any external data is unavailable on the server, so we can only build a balanced data via clients data.
% we decouple the global model learning procedure.
% that is, after training the fedavg model, we fix the feature extractor and only re-train the classifier. We conduct a quick experiment to re-train the global model classifier on a balanced dataset collected from clients. 
%In this work, we address the joint problem of the heterogeneous and long-tailed data distribution in federated learning from a novel perspective based on an intuitive observation: heterogeneous and long-tailed distribution might not be an issue in learning high-quality representations, and the biased classifier is the true devil leading to the poor performance of the global model. 
%Motivated by the discovery, we conduct a quick experiment to adjust the FedAvg model via re-training the classifier on a balanced dataset collected from clients. 
Given a global model pre-trained on heterogeneous and long-tailed data, we fix its feature extractor and re-train its classifier on the server with a small balanced dataset (100 per class) collected from clients. 
It can be observed in Figure \ref{motivation_figure} that the strategy is particularly useful and the classification performance is significantly improved.
%with only a small balanced dataset. %which indicates that heterogeneous and long-tailed distribution might not be an issue in learning high-quality representations.
In particular, re-training classifier achieves the highest performance gain of 18\% when $\IF=100$ and $\alpha=0.1$ (the rightmost grey bar in the left subfigure of Figure \ref{motivation_figure}), which is an extremely heterogeneous and long-tailed setting.
% which is consistent with our analysis that re-training the classifier works best in highly heterogeneous and long-tailed setting.
However, this prerequisite of using a shared balanced dataset on the server is infeasible for most FL practical scenarios due to privacy concerns. %a specially collected dataset may not be available on the server. As shown in the above experiment, sharing data between server and client leads to privacy leaks. However, the data collected by the server itself may not follow the same distribution as the training data.

Motivated by the idea of two-stage learning and the privacy concerns of FL, we propose a novel and privacy-preserving FL method called Classifier Re-training with Federated Features (CReFF) to deal with the joint problem of data heterogeneity and long-tail distribution. 
We learn a set of balanced features called \textit{federated features} because only the classifier needs to be re-trained on the server.
%It is based on an intuitive idea that the classifiers re-trained on real and federated features should converge to a similar solution in the parameter space, which will produce approximate performance. In other words, the federated features should make the optimization of classifier re-training follow a similar path as re-training on the real data. Therefore, we optimize theavailable federated features to make their gradients close to the gradients of real data. 
It is based on an intuitive idea that the classifier re-trained on federated features should produce comparable performance as the one re-trained on the real data, which can be achieved by making two classifiers similar.
Therefore, the classifier re-training optimization on federated features should follow the similar path as that of the real data. Specifically, we optimize the federated features to make their gradients close to the gradients of real data. 
As shown in Figure \ref{motivation_figure}, the classifier re-trained on federated features (orange bars) achieves comparable performance as the one re-trained on real data (grey bars). Extensive experiments show that CReFF significantly outperforms the state-of-the-art federated learning methods on image classification tasks with heterogeneous and long-tailed data distribution.
The contributions of this paper can be summarized as follows:
\begin{itemize}
	\item We study the joint problem of FL with heterogeneous and long-tailed data distribution, where the local and global class distribution is unknown to the server. 
	\item We reveal an intriguing fact that the biased classifier is the primary factor leading to the poor performance of an FL global model on heterogeneous and long-tailed data.
	\item We propose CReFF, a novel FL algorithm to deal with heterogeneous and long-tailed data by re-training the classifier with learnable federated features on the server. CReFF has no  privacy concerns because no real data or information of class distribution is required. 
\end{itemize}
\section{Related Work}
%% FL中解决Non-iid的方法
\subsection{Federated Learning with Heterogeneous Data}
%%许多方法去解决non-iid问题，但是几乎没有解决LT问题
A variety of solutions have been proposed to tackle data heterogeneity in FL, 
%which are roughly categorized into client-side methods and server-side methods. The former focuses on regularizing the objective of local training process such that the diversity of client models can be limited \cite{MLSYS2020_38af8613,huang2021behavior}, and the later adopt specific optimization \cite{NEURIPS2020_18df51b9,chen2021fedbe} to alleviate the negative influence of data heterogeneity.
which are mainly from two perspectives: One focuses on optimization strategies to make the diversity between client models and global model limited \cite{MLSYS2020_38af8613,huang2021behavior}; the other adopt mechanisms on the server to alleviate the negative influence of data heterogeneity \cite{NEURIPS2020_18df51b9,chen2021fedbe}. 
%regularizing the objective of local training process such that the diversity of client models can be limited \cite{MLSYS2020_38af8613,NEURIPS2020_564127c0}; the other adopt specific optimization to alleviate the negative influence of data heterogeneity \cite{NEURIPS2020_18df51b9,huang2021behavior}. 
%One focuses on regularizing the objective of local training process such that the diversity of client models can be limited \cite{MLSYS2020_38af8613,NEURIPS2020_564127c0}; the other adopt specific optimization to alleviate the negative influence of data heterogeneity \cite{NEURIPS2020_18df51b9,huang2021behavior}. 
For example, CCVR \cite{2021No} deals with data heterogeneity via classifier re-training using virtual features sampled from an approximated Gaussian mixture model.
Although the abovementioned methods solve the data heterogeneity to some extent, they poorly perform on the tail classes due to the lack of consideration of the global long-tail distribution. 

Some FL methods are designed for imbalanced data (not specifically long-tailed). One strategy is to adopt client selection to match clients with complementary class distribution \cite{duan2020self,yang2020federated}. In this case, all clients are required to upload their local class distributions to the server, which violates the principle of privacy protection of FL.
Ratio loss \cite{wang2021addressing} utilizes balanced auxiliary data on the server to estimate the global class distribution for better local optimization. However, as we mentioned above, the auxiliary data is not available on the server in real applications. Compared with these related methods, the proposed CReFF has no privacy concerns because no real data or information of class distribution is required. 

%%LT中的经典方法：Decouple(重训练分类器)
\subsection{Long-tail learning}
%Recent studies of long-tail learning can be categorized into the following three directions: 
Recently, long-tail learning has drawn much interest in deep learning \cite{2021deeplt}. Some methods follow the ideas of imbalance learning to augment the feature space for rare classes \cite{kim2020m2m,zang2021fasa} or re-weight different classes according to their frequencies \cite{NEURIPS2019_621461af}.
%to achieve a more balanced data distribution \cite{chou2020remix,kim2020m2m}, or assign different weights to training samples from different classes \cite{NEURIPS2019_621461af,cui2019class}.
Some recent methods decouple the training phase into representation learning and classifier re-training \cite{Kang2020Decoupling}, which generate more generalizable representations and achieve strong performance after re-balancing the classifier. Based on decoupling, ensemble-based methods \cite{xiang2020learning,Xudong2021Longtailed} adopt a multi-expert framework to learn diverse classifiers in parallel, which reduces the model bias towards the tail classes. However, most of them require the global class distribution, as we mentioned above. During FL model training, it is infeasible to gather the information of class distribution of each client to obtain the global class distribution, which makes the vast majority of long-tail learning methods not applicable to FL scenarios.
%%%%%%%%%%%%%%%%%%%%%%%%%%%%%%%%%%%%%%%%
\section{Proposed Method}
\subsection{Preliminaries}
%We now introduce the problem setting of federated learning. Subsequently, we perform an empirical study to explore the impact of non-IID and long-tail distribution on the global model, which motivates our proposed approach.
\textbf{Settings and Notations.} We discuss a typical FL setting with $K$ clients holding potentially heterogeneous data partition $\DD^1, \DD^2, ..., \DD^K$, respectively. The goal is to learn a global model over the whole training data $\DD\triangleq \bigcup_{k} \DD^k$ with the help of a central server without data transmitting. 
In this paper, we consider the setting when $\DD$ is drawn from a long-tail distribution $\mathcal{X}=\{\bx_i, y_i\}|_{i=1}^N, y_i\in\{1,...,C\}$.
Let $n_c^k$ be the number of samples of class $c$ on client $k$, and $n_c=\sum_{k=1}^Kn_c^k$.
%where $\sum_{j=1}^{n_j}=N$. 
%Let $n_j^k$ be the number of global training sample for class $j$, where $\sum_{j=1}^{n_j}=N$. 
Without loss of generality, a common assumption in long-tail learning is that the classes are sorted by cardinality in non-increasing order, i.e., if $c_1 < c_2$, we have $n_{c_1} \geq n_{c_2}$ and $n_1\gg n_C$. 
%The imbalance factor ($\IF$) is defined as $n_1/n_C$.
%we assume that the size of data of all classes obeys the power-law and the classes are sorted by cardinality in decreasing order and , i.e., if $i < j$, then $N_i > N_j$.
%$X=\{\bx_i, y_i\},i\in\{1,..,N\}$.  Without loss of generality, we assume that the classes are sorted by cardinality in decreasing order, i.e., if $i < j$, then $N_i \gg N_j$.
%$(\mathcal{X},\mathcal{Y})$, $\mathcal{Y}\in\{1,...,C\}$. 
For an FL model, we typically consider a neural network $\phi_{\bw}$ with parameters $\bw=\{\bu, \bv\}$. It has two main components: 1) a feature extractor $f_{\bu}$ with parameters $\bu$, mapping each input sample $\bx$ to a $d$-dim feature vector; 2) a classifier $h_{\bv}$ with parameters $\bv$, typically being a fully-connected layer which outputs logits to denote class confidence scores. The parameters of the $k$th client local model are denoted as $\bw_k$.
%\subsection{Basic Algorithm of Federated Learning}\label{fl}

\noindent\textbf{Basic Algorithm of Federated Learning.} FedAvg \cite{mcmahan2017communication} is the fundamental algorithm for FL. In round $t$, the server first sends a global model $\bw^t$ to clients. The clients then update the received model on their local data $\DD^k$, $k=1,...,K$:
\begin{align}\label{client_update}
\bw_k^{t+1}\gets \bw_k^t-\eta\nabla_{\bw}\ell(\bw^t;\DD^k).
\end{align}
%: $\min_{\bw_k^{t+1}}\mathbb{E}_{(\bx_i, y_i)\sim\DD^k}\ell(\bw_k^t;(\bx_i, y_i))$. 
After local updating, some clients in the set $\mathcal{A}^t$ are selected to upload their updated models to the server. Finally, the server performs weighted average to update the global model for round $t+1$:
\begin{align}\label{global_agg}
%\bw^{t+1}&= \sum_{k=1}^{\left| \mathcal{A}^t \right|}\frac{|\DD^k|}{|\DD|}\bw_k^t
\bw^{t+1}&= \sum_{k\in \mathcal{A}^t}\frac{|\DD^k|}{\sum_{k\in\mathcal{A}^{t}}|\DD^k|}\bw_k^{t+1}.
\end{align}
%where, $\left| \mathcal{A}^t \right|$ is the number of selected clients in the $t$th round.

\subsection{Motivation}
%\subsection{Motivation}
%it can be found that heterogeneous and long-tailed data distribution might not be an issue in learning high-quality representations.
%%用少量数据重训练分类器可以极大程度的缓解长尾和异构带来的影响。
Classifier re-training has been shown effective for heterogeneous data \cite{2021No} and long-tailed data \cite{Kang2020Decoupling}, separately. Now, we empirically show that classifier re-training also works for the joint problem. In other words, the biased classifier is the primary factor leading to the poor performance of the global model trained on heterogeneous and long-tailed data.
%heterogeneous and long-tailed distribution might not be an issue in learning high-quality representations, 
% we perform a thorough experimental investigation on each layer of a deep network.
%explore the impact of heterogeneous and long-tailed data distribution on the each component of a deep network.
%our assumption that heterogeneous and long-tailed distribution might not be an issue in learning high-quality representations, and the biased classifier is the true devil, 
%%在少量数据前提下
%In order to further verify our assumption, we first perform a thorough experimental investigation on each component of a ResNet-8 network under a setting of heterogeneity and long-tail distribution. 

\begin{figure}[!t]
	\centering
	\includegraphics[width=.9\linewidth]{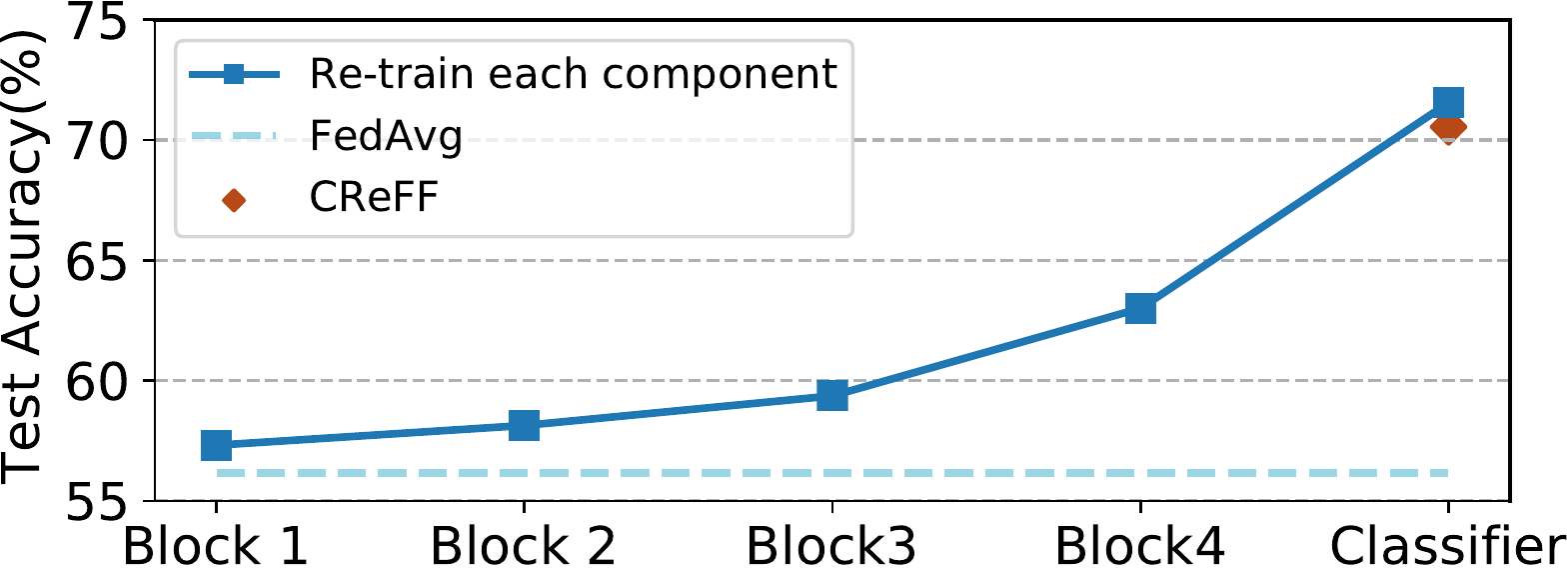}
	\caption{The performance of re-training each component of a FedAvg model pre-trained on CIFAR-10-LT with $\IF=100$ and $\alpha=0.5$. 
	}
	\label{Re_train_layer}
\end{figure}

First, we divide a ResNet-8 network into five components, including four blocks\footnote{We regard the first convolutional layer with BN layer as the first block.} and one classifier. 
Then, we use a small set of balanced data (100 per class) collected from clients to re-train each component of a FedAvg model while keeping other components fixed. The FedAvg model is pre-trained on CIFAR-10-LT with $\IF=100$ and $\alpha=0.5$. 
%%%为什么在serve端只能有少量数据？
%Note that in the FL environment, any data is unavailable on the server due to privacy concerns and computation burden. In this experiment, we can only fetch a small balanced labeled data from clients like previous studies \cite{wang2021addressing,2021No}.
%there is only a small amount of or even no labeled data on the server due to privacy concerns \cite{wang2021addressing,chen2021fedbe}
As shown in Figure \ref{Re_train_layer}, we can observe that re-training any single component can achieve certain improvement (blue line). Especially, re-training the classifier achieves the highest performance gain (around 15.35\%). 
%We also validate that whether more performance gain can be achieved when more components are re-trained. We re-train the rest of network starting from a given component, as shown in Figure 2 (orange line).
%%So we wonder if more performance gain could be obtained by re-training more components, which motivates us to conduct the second experiment. 
%%%the rest 有问题?
%For example, if we re-train from block 3, all the parameters in block 3, block 4, and the classifier are re-trained. With only a small set data, it can be observed that the more components are re-trained, the model will perform worse, which shows that only a small data is not capable enough to re-train a large network. However, a large balanced data is unavailable in practice, as we mentioned above.
%
%
%re-training a single block performs better and re-training the classifier achieves the best performance. The reason why re-training more components becomes worse is that only a small data is not capable enough to re-train a large network. %However, a large balanced data is unavailable on the server.
Therefore, it can be shown that the biased classifier is the primary factor leading to the poor performance of the global model because other components are less affected.
% with heterogeneous and long-tailed data.
% because other components are less affected.
%However, re-training the classifier only using the small data can also achieve the high performance. 
%The observation is thought-provoking: re-training the classifier achieves the highest performance gain (green line), which shows that the biased classifier is the primary reason leading to the poor performance of the global model. 
%% 重新训练多了

However, requiring external data on the server is impractical in real FL applications, although the amount of the data is small. Transmitting data from clients violates the key privacy-preserving principle of FL, and the data collected by the server itself may not follow the same distribution as the training data in clients. 
%%% 强调区别于CCVR
%%%%%%%%%%%%%%%%%%%%%%%%%%%%%%%%%%%%%%%%
\subsection{Framework of the proposed CReFF}

\begin{figure*}[!t]
	\centering
	\includegraphics[width=\linewidth]{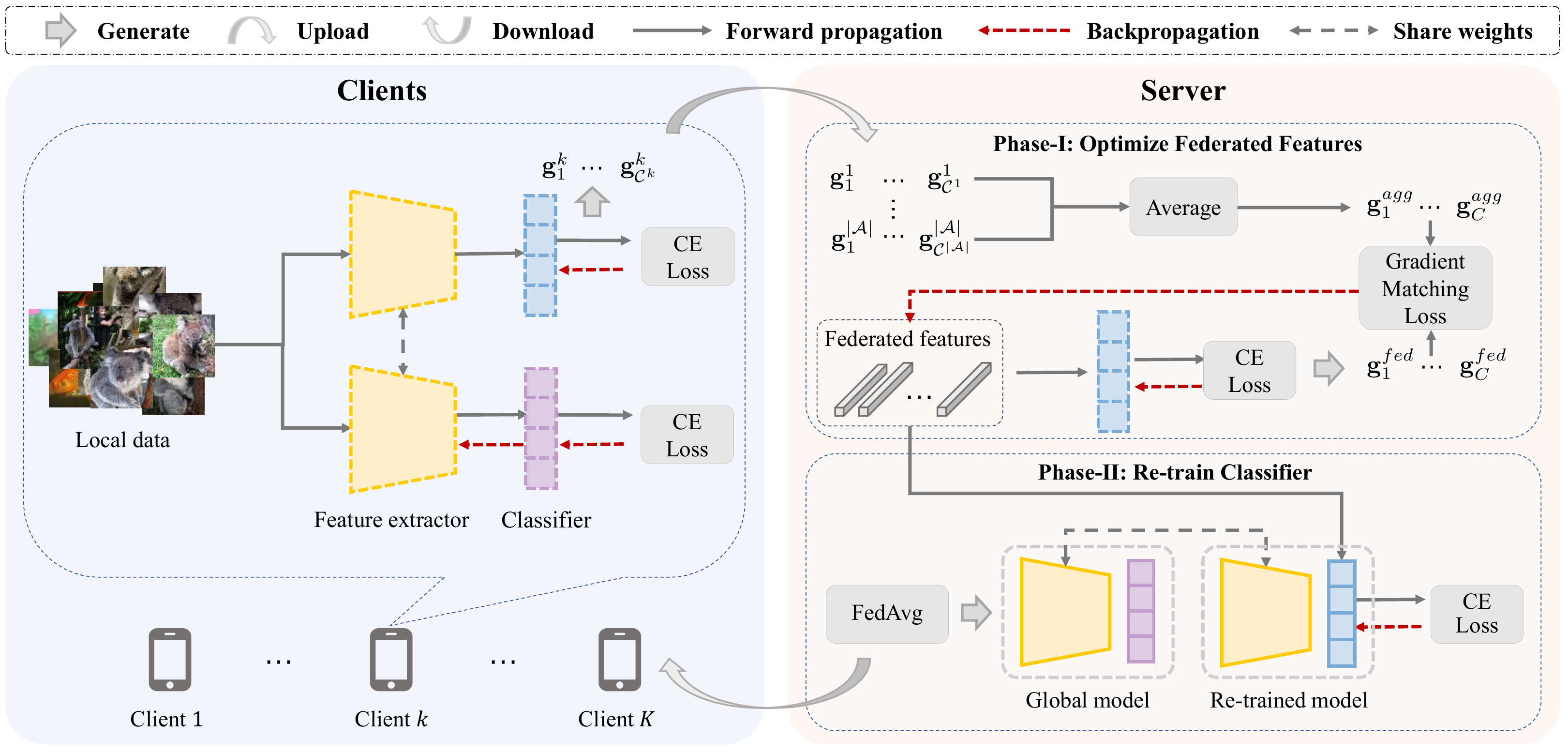}
	\caption{The framework of CReFF. In each round, clients send updated local models and real feature gradients to the server, and the server sends the aggregated global model and the re-trained model to clients.}
	\label{model_figure}
	%\vspace{-2px}
\end{figure*}
%%方法分为两部分：global model的feature extractor用FedAvg的方法进行训练(parameter averaging); classifier用生成feature re-training，并且下一个round用得到的balanced classifier训练得到feature
Based on the above assumption and corresponding observation, we propose CReFF to deal with the joint problem of data heterogeneity and long-tail distribution. It is designed as a simple and effective approach based on FedAvg by only re-training a new classifier using a small set of learnable features on the server, which is called \textit{federated features}. 
%CReFF has no privacy concerns because any external data and distribution information are not acquired.
%With federated features, no real data is needed from any client in the process of re-training.
%The goal is to synthetic a small set of balanced features with their labels $\DS=(\bs_i, y_i)|_{i=1}^M$, where $\bs\in\mathbb{R}^d$, $y\in\{1,...C\}$. Let $$
%, and $M\ll N$, and then an unbiased classifier can be re-trained on them.
%%%逻辑好像不是很通
%We learn the federated features based on an intuitive idea that the classifiers re-trained on federated and real features should converge to a similar solution in the parameter space, which will produce approximate performance.
Our goal is to re-train the classifier on federated features such that its performance is comparable with the classifier re-trained on real features, which can be achieved by making two classifiers similar.
Therefore, the classifier re-training optimization on federated features should follow the similar path as that of the real data.
Specifically, we optimize the federated features to make their gradients close to the gradients of real data such that the classifiers trained on real and federated features converge to a similar solution in the parameter space. 
In this way, the classifier re-trained on the federated features can be approximated by the one re-trained on the real data in a privacy-preserving manner. As shown in Figure \ref{Re_train_layer}, CReFF produces comparable performance as on real data.
%In words, we wish that the classifier parameters trained on federated features are similar to the ones trained on the real features at each round. It also means that these federated features contain the meaningful information derived from the local training data, so an unbiased classifier can be trained on them.
%It also means that these synthetic features contain the meaningful information derived from the local training data, so an unbiased classifier can be trained on them.
%Inspired by DC \cite{2020Dataset}, we adopt gradient matching 
%In order to synthesize high-quality features, we encapsulate the knowledge of the local training data into a small number of synthetic features on the server. 
%%压缩了client的样本之后,我们有了合成特征的能力。然后我们就可以在server段生成平衡的特征来得到unbiased classifier。
%These synthetic features contain the meaningful information derived from the local training data, so an unbiased classifier can be trained on them.
%These realistic features contain the meaningful information derived from the local training data, so an unbiased classifier can be trained on them. XXX can achieve significant performance gains under all the settings, as shown in Fig. \ref{motivation_figure} (b). 

%%描述整个思路
CReFF consists of two core components: local training on clients and federated features optimization on the server. In round $t$, each client receives two models from the server: 
%1) a FedAvg model $\bw^t$ composed of a feature extractor $\bu^t$ and a classifier $\bv^t$; 2) an re-trained model $\hat{\bw}^t$ composed of a feature extractor $f\bu^t$ and a re-trained classifier $\hat{\bv}^t$. The former is used to update client model, and the latter is used to calculate the gradient w.r.t. $\hat{\bv}^t$ to make the optimization path similar. After local training, some clients in the set $\mathcal{A}$ are selected to upload two corresponding parts to the server. Finally, the server performs weight average to update the FedAvg model and the re-trained model for round $t+1$.
1) a global model $\bw^t$; 2) a re-trained model $\widehat{\bw}^t$. The former is used to update local models for global model aggregation, and the latter is used to calculate the real feature gradients for federated features optimization. After local training, some clients are selected to upload their local models and real feature gradients to the server. 
Finally, the server updates the global model and the re-trained model for round $t+1$. The architecture of CReFF is illustrated in Figure \ref{model_figure}.
%%为了获取local data的知识，并且把知识迁移给server端的生成feature
%We adopt gradient matching to transfer the knowledge derived from the local training data to the synthetic features on the server-side. 
% In each round of XXX
%Nevertheless, in our method, clients only upload the average gradients of each class to the server, which erases the private information of local data to some context because the averaging operation is irreversible.

\noindent\textbf{Local Training.} 
%As shown in Fig. \ref{model_figure}
In round $t$, the server sends two models to clients. The global model $\bw^t$ is composed of a feature extractor $\bu^t$ and a classifier $\bv^t$, and the re-trained model $\widehat{\bw}^t$ has a re-trained classifier $\widehat{\bv}^t$ with the same feature extractor $\bu^t$. For each received model, the local training consists of a corresponding part. The first part is a typical FedAvg local model update as shown in Equation (\ref{client_update}).
%First, client $k$ updates the FedAvg model $\bw_k^t$ on local data $\DD^K$,
%\begin{align}\label{local_update}
%\bw_k^{t+1}\gets \bw_k^t-\eta\nabla_{\bw_k^t}\ell(\bw_k^t;(\bx_i, y_i))
%%\min_{\bw_k^t}\mathbb{E}_{(\bx_i, y_i)\sim\DD^k}\ell(\bw_k^t;(\bx_i, y_i))
%\end{align}
The second part is calculating the real feature gradients for federated features optimization on the server. %The goal is to synthetic a small set of balanced features with their labels $\DS=(\bs_i, y_i)|_{i=1}^M$, where $\bs\in\mathbb{R}^d$, $y\in\{1,...C\}$. Let $$
%, and $M\ll N$, and then an unbiased classifier can be re-trained on them.
Client $k$ produces $d$-dim real features $\mathcal{Z}_c^k=\{\bz_{c,i}^k\}|_{i=1}^{n_c^k}$ for class $c$ through the feature extractor $\bu^t$. Then, real feature gradients $\bg_c^k\in\mathbb{R}^{C\times d}$ of class $c$ can be computed using $\widehat{\bv}^t$:
%by using the received re-trained classifier $\widehat{\bv}^t$ 
%with a given loss function $\ell$:
\begin{align}\label{local_gradient}
\bg_c^k &= \frac{1}{n_{c}^k}\sum_{i=1}^{n_c^k}\nabla_{\widehat{\bv}}\ell(h_{\widehat{\bv}^t}(\bz_{c,i}^k), y_i).
\end{align}
% Note that the reason why we compute local gradient on re-trained classifier instead of global classifier is that the latter is biased, causing the gradient to be also biased, which cannot provide a helpful guide to the synthetic features on the tail classes.
Finally, client $k$ uploads two parts to the server: 1) the local model $\bw_k^{t+1}$ for global model aggregation; 2) the real feature gradients $\{\bg_c^k|c\in\CC^k\}$ for federated features optimization. Herein, $\CC^k$ denotes the set of classes on client $k$ because the real feature gradients on client $k$ may only contain partial classes due to data heterogeneity. It is worth noting that clients only upload the average gradients of each class to the server, which erases the private information of local data to some context because the averaging operation is irreversible.

\noindent\textbf{Federated features optimization.} 
Once the server receives two parts from the selected clients in $\mathcal{A}^t$, there are also two corresponding parts to aggregate the client models and optimize federated features for classifier re-training, respectively. The first part is to aggregate the local models by Equation (\ref{global_agg}) to obtain the aggregated global model $\bw^{t+1}$. The second part aims to learn a set of balanced $d$-dim federated features $\DS_c^t=\{\bs_{c,i}^t\}|_{i=1}^{m}$ for each class $c$ in round $t$, where $m$ is the number of federated features of each class.
%with their labels $\DS=(\bs_i, y_i)|_{i=1}^M$, where $\bs\in\mathbb{R}^d$, $y\in\{1,...C\}$. Let $m_c$ is the number of features of class $c$, and $M=\sum_{c=1}^Cm_c$.
%Let $m_c$ be the number of samples of class $c$, and $\sum_{c=1}^Cm_c=M$. 
First, the server aggregates real feature gradients of each class $c$ by averaging over all selected clients in $\mathcal{A}^t$:
\begin{align} \label{aggre_gradient}
\bg^{agg}_c &= \frac{1}{\left| \mathcal{A}_c^t\right|}\sum_{k=1}^{|\mathcal{A}_c^t|}\bg_c^k,
\end{align}
where $\mathcal{A}_c^t$ is the subset of clients that hold class $c$. 
%where $\left| \mathcal{A}^t\right|$ is the number of selected clients on round $t$. 
%our goal is to learn a small set of balanced federated features 
Then, the server computes the federated feature gradients $\bg_c^{fed}$ over federated features $\DS^t$ of class $c$ by using the same re-trained classifier $\widehat{\bv}^t$ as calculating $\bg^{agg}_c$:
\begin{align}\label{federated_gradient}
\bg_c^{fed}&=\frac{1}{m}\sum_{i=1}^{m}\nabla_{\widehat{\bv}}\ell(h_{\widehat{\bv}^t}(\bs_{c, i}^t), y_i).
\end{align}
Both $\bg^{fed}_c$ and $\bg_c^{agg}$ are gradients of the same re-trained classifier $\widehat{\bv}$, whose dimensions are $C\times d$. We use the gradient matching loss \cite{zhao2020dataset} to measure the difference between them by averaging the cosine dissimilarity of each part:
\begin{align}\label{cosine}
%D(\bg^{fed}_c, \bg^{agg}_c) = 1-\frac{F(\bg^{fed}_c)\cdot F(\bg^{agg}_c)}{\|F(\bg^{fed}_c)\|\|F(\bg^{agg}_c)\|}
D(\bg^{fed}_c, \bg^{agg}_c) =\frac{1}{C}\sum_{j=1}^C \big(1-\frac{\bg^{fed}_c[j]\cdot \bg^{agg}_c[j]}{\|\bg^{fed}_c[j]\|\|\bg^{agg}_c[j]\|}\big),
\end{align}
where $\bg[j]$ denotes the $j$th row of the gradient. After optimizing the federated features $\DS^t$ on this loss, we obtain an updated $\DS^{t+1}$ for classifier re-training. 

Compared with sampling virtual features from the approximated Gaussian mixture for classifier re-training \cite{2021No}, CReFF learns federated features by optimizing the gradient matching loss. 
The advantages are twofold. On the one hand, CReFF can produce nearly the same gradients as the real features to update the classifier, which can not be achieved by CCVR. On the other hand, CCVR requires each client to upload its local class distribution to the server for computing global mean and covariance. In contrast, CReFF does not require the information of local class distribution in a more privacy-preserving manner.

The last step of CReFF is to re-train the classifier of the updated global model $\bw^{t+1}$ to obtain the re-trained model $\widehat{\bw}^{t+1}$. Initialized as $\widetilde{\bv}$, a new classifier $\widehat{\bv}^{t+1}$ is re-trained on the optimized balanced federated features $\DS^{t+1}$:
\begin{align}\label{re-train}
\widehat{\bv}^{t+1}\gets \widetilde{\bv} - \eta\nabla_{\widetilde{\bv}}\ell(\widetilde{\bv};(\bs_i^t, y_i)).
\end{align}
Finally, the updated global model $\bw^{t+1}=\{\bu^{t+1}, \bv^{t+1}\}$ and the re-trained model $\widehat{\bw}^{t+1}=\{\bu^{t+1}, \widehat{\bv}^{t+1}\}$ are sent to clients for round $t+1$. The whole training process of CReFF is shown in Algorithm 1.

\begin{algorithm}[tb]
	\caption{Training process of CReFF}
	\label{alg:algorithm}
	\LinesNumbered %要求显示行号
	\KwIn{Initialized global model $\bw^0$, number of steps $I$ for federated features optimization, number of steps $J$ for classifier re-training, number of federated features per class $m$, number of training rounds $T$}%输入参数
	%%%像是转置
	\KwOut{Re-trained model $\widehat{\bw}^T$ on round $T$ }%输出
	\For{$t=1$ \KwTo $T$}{
		%\tcp{Server executes:}
		Randomly select a set of active clients $\mathcal{A}^t$;\\
		\tcp{Clients execute:}
		\For{$k\in \mathcal{A}^t$}{	
			Update local model $\bw_k^{t+1}$ by Equation (\ref{client_update});\\
			Compute real feature gradients $\{\bg_c^k|c\in\CC^k\}$ by Equation (\ref{local_gradient});\\
			Send $\bw_k^{t+1}$ and $\{\bg_c^k|c\in\CC^k\}$ to the server;\\
		}
		\tcp{Server executes:}
		Aggregate local models to $\bw^{t+1}$ by Equation (\ref{global_agg});\\
		Aggregate real feature gradients to $\bg^{agg}_c$ by Equation (\ref{aggre_gradient});\\
		Compute federated feature gradients $\bg^{fed}_c$ by Equation (\ref{federated_gradient});\\
		%		Compute federated features gradients $\bg_c^{fed}$ by  Equation (\ref{federated_gradient});\\
		Optimize federated features to $\DS^{t+1}$ with the loss in Equation (\ref{cosine}) for $I$ epochs;\\
		Re-train the classifier of $\bw^{t+1}$ to $\widehat{\bw}^{t+1}$ on $\DS^{t+1}$ by Equation (\ref{re-train}) for $J$ epochs;\\
		%	  $\widehat{\bw}^{t+1}=\{\bu^{t+1}, \widehat{\bv}^{t+1}\}$\\
		Send $\bw^{t+1}$ and $\widehat{\bw}^{t+1}$ to clients;\\
		%Send $\bw^{t+1}=\{\bu^{t+1}, \bv^{t+1}\}$ and $\widehat{\bw}^{t+1}=\{\bu^{t+1}, \widehat{\bv}^{t+1}\}$ to clients;\\
	}
\end{algorithm}
\section{Experimental Results}
In this section, we first show the ability of CReFF to deal with data heterogeneity and long-tail distribution on several benchmark datasets, compared with the state-of-the-art FL methods. Then, we specifically study the reason of federated features being effective for classifier re-training on the server.
\subsection{Experimental Setup} \label{setting}
% 用bullets把数据写两小段
We conduct experiments on the following datasets:

% 用bullets把数据写两小段
$\bullet$ CIFAR-10/100-LT \cite{2009Learning}. We follow \cite{NEURIPS2019_621461af} to shape the original balanced CIFAR-10/100 into long-tail distribution with $\IF=100$, 50 and 10, respectively. Like previous studies \cite{NEURIPS2020_18df51b9}, we use Dirichlet distribution to generate the heterogeneous data partition among clients. We set the value of $\alpha$ at 0.5 on CIFAR-10/100-LT. 

$\bullet$ ImageNet-LT \cite{2015ImageNet}. It contains 115.8K images from 1,000 categories, with the largest and smallest categories containing 1,280 and 5 images, respectively. We set the value of $\alpha$ at 0.1 on ImageNet-LT. 
%It is constructed from ImageNet by sampling a subset following the Pareto distribution \cite{liu2019large}. 

%$\bullet$iNaturalist 2018 \cite{Horn_2018_CVPR} is a classification dataset, which is on a large scale and suffers from extremely imbalanced label distribution. It is composed of 437.5K images from 8,142 categories.

% 网络
We use ResNet-8 for CIFAR-10/100-LT, and ResNet-50 for ImageNet-LT as the base model. We implement all compared FL methods with the same model for a fair comparison.
%编程语言与参数
All experiments are run by PyTorch on two NVIDIA GeForce RTX 3080 GPUs. By default, we use standard cross-entropy loss and run 200 communication rounds. We set the number of total clients at 20 and an active client ratio $40\%$ in each round. For local training, the batch size is set at 32. For server-side training, we initialize the federated features as random noise and set the number of them per class at 100, the optimization steps $I$ on federated features at 100, the classifier re-training steps $J$ at 300. We use SGD with a learning rate $0.1$ as the optimizer for all optimization process.

%%ImageNet中B=128, alpha=0.1
%数据异质

\begin{table*}[!t]
	\centering
	\begin{tabular}{@{}llcccccc@{}}
		\toprule
		&                    & \multicolumn{3}{c}{CIFAR-10-LT}                                & \multicolumn{3}{c}{CIFAR-100-LT}                               \\ \cmidrule(l){3-8} 
		\multirow{-2}{*}{\textbf{Family}}        & \multirow{-2}{*}{\textbf{Method}}   & \multicolumn{1}{c}{$\IF=100$}    & \multicolumn{1}{c}{$\IF=50$}    & \multicolumn{1}{c}{$\IF=10$}    & \multicolumn{1}{c}{$\IF=100$}    & \multicolumn{1}{c}{$\IF=50$}    & \multicolumn{1}{c}{$\IF=10$}    \\ \midrule
		& FedAvg                 & 56.17             & 59.36             & 77.45             & 30.34             & 36.35             & 45.87             \\
		& FedAvgM                & 52.03             & 57.11             &   70.81         &       30.80      &    35.33       &   44.66      \\
		&FedProx  &     56.92         &   60.89          &   76.53           &   31.67         &    36.30         & 46.10 \\
		& FedDF                 &     55.15        &  58.74            &   76.51        &  31.43          &         36.22                 &   46.19                      \\
		& FedBE                 &   55.79                  & 59.55
		& 77.78                      &   31.97                  &   36.39    
		& 46.25 \\ 
		& CCVR                 &   69.53                  & 71.89
		& 78.48                      &   33.43                  &   36.98    
		& 46.88 \\ 
		\multirow{-8}{*}{\begin{tabular}[c]{@{}l@{}}Heterogeneity-oriented\\ FL methods\end{tabular}} 
		& FedNova                &   57.79          &     63.91         &  77.79          &     32.64        &    36.62         &   46.75          \\	\midrule
		& Fed-Focal Loss           &   53.83           &     57.42      &     73.74      &   30.67        &    35.25         &   45.52           \\
		& Ratio Loss         &     59.75      
		&  64.77
		&   78.14         
		&32.95
		&  36.88
		& 46.79
		\\   
		\multirow{-3}{*}{\begin{tabular}[c]{@{}l@{}}Imbalance-oriented\\ FL methods\end{tabular}}
	   	& FedAvg+$\tau$-norm           & 49.95
	   & 51.41
	   & 72.08
	   & 26.22
	   & 33.71
	   & 43.65 	\\      \midrule
		\multirow{-1}{*}{Proposed method}
		& \cellcolor[HTML]{EFEFEF}CReFF & \cellcolor[HTML]{EFEFEF}\textbf{70.55} & \cellcolor[HTML]{EFEFEF}\textbf{73.08} & \cellcolor[HTML]{EFEFEF}\textbf{80.71} & \cellcolor[HTML]{EFEFEF}\textbf{34.67} & \cellcolor[HTML]{EFEFEF}\textbf{37.64} & \cellcolor[HTML]{EFEFEF}\textbf{47.08}\\\midrule
		%		\\ \cmidrule(l){2-8} 
		%		& FedProx+XXX &70.52  & 72.09 & &  &  &  \\ 
		%		& FedNova+XXX & 70.92 & 73.05 & &  &  &\\	
		
		%	& XXX &  &  & &  &  &\\ \midrule
	\end{tabular}
	\caption{Top-1 test accuracy ($\%$) achieved by compared FL methods and CReFF on CIFAR-10/100-LT with different $\IF$s.}
	\label{t1}
\end{table*}

%\setlength{\textfloatsep}{5pt}
%%初始化的$\bw$要和$\hat{\bw}$的特征提取器相同
\subsection{Comparison with the State-of-the-art Methods}
We compare CReFF with several heterogeneity-oriented FL methods, including FedAvg \cite{mcmahan2017communication}, FedAvgM \cite{hsu2019measuring}, FedProx \cite{MLSYS2020_38af8613}, FedDF \cite{NEURIPS2020_18df51b9}, FedBE \cite{chen2021fedbe}, CCVR \cite{2021No} and FedNova \cite{NEURIPS2020_564127c0}. Moreover, we also compare with imbalance-oriented FL methods, including Fed-Focal Loss \cite{sarkar2020fed}, Ratio Loss \cite{wang2021addressing} and FedAvg with $\tau$-norm \cite{Kang2020Decoupling}. Note that Ratio Loss depends on a balanced auxiliary data, so we fetch the data from clients, although it is infeasible in practice.
\begin{table}[!t]
	\centering
	\begin{tabular}{@{}llccc@{}}
		\toprule
		& \multicolumn{4}{c}{ImageNet-LT} \\ \cmidrule(l){2-5} 
		\multirow{-2}{*}{\textbf{Method}} & All  & Many  & Medium & Few \\ \midrule
		FedAvg             &  23.85  &  34.92  &  19.18   &  7.10 \\
		FedAvgM              &  22.57  &  33.93  &  18.55   & 6.73  \\
		FedProx        &22.99      & 34.25   &  17.06  &     6.37   \\
		FedDF               & 21.63   &  31.78  &  15.52   &  4.48 \\
		CCVR               &  25.49  &36.72    &20.24     &9.26   \\
		%\midrule
		Fed-Focal Loss        &  21.60    &  31.74  & 15.77   &  5.52      \\
		Ratio Loss            &  24.31  &   36.33 &   18.14  &  7.41 \\
		FedAvg+$\tau$-norm                &  21.58  &  31.66  &  15.76   & 4.33  \\
		\midrule
		\rowcolor[HTML]{EFEFEF} 
		CReFF        &  \textbf{26.31}  &  \textbf{37.44}  &   \textbf{21.87} & \textbf{10.29}  \\ \bottomrule
	\end{tabular}
	\caption{Top-1 test accuracy ($\%$) achieved by compared FL methods and CReFF on ImageNet-LT.}
	\label{t2}
\end{table}

\noindent\textbf{Results on CIFAR-10/100-LT.} 
%%在CIFAR100上提升比较少，是因为其特征提取器就不太好;
The results are summarized in Table \ref{t1}. CReFF achieves the highest test accuracy on both datasets with different $\IF$s. Compared with the baseline FedAvg, the performance gain of CReFF is the highest when $\IF=100$ (around 14.4\% improvement for CIFAR-10-LT and 4.3\% improvement for CIFAR-100-LT). It shows the generalization ability of CReFF when the global class distribution is highly long-tailed. 
%Moreover, we observe that the improvement on CIFAR-100-LT seems subtle compared with that of CIFAR-10-LT. 
For heterogeneity-oriented methods, most of them (e.g., FedAvgM and FedDF) perform similarly to FedAvg because they only deal with data heterogeneity without taking global class imbalance into account. 
For imbalance-oriented methods, some of them, e.g., Ratio Loss, perform well in some cases compared with FedAvg. However, there is still a performance gap compared with CReFF because they aim at dealing with the global imbalance problem but generally ignore the data heterogeneity problem across clients.

\noindent\textbf{Results on ImageNet-LT.} 
We further evaluate CReFF on ImageNet-LT as reported in Table \ref{t2}. To better examine the performance of classes with different number of training samples, we report the accuracy on three sets of classes: Many-shot (more than 100 samples), medium-shot (20-100 samples), and few-shot (less than 20 samples). 
Compared with other FL methods, CReFF achieves the best results on all cases, which shows that the improved overall accuracy by CReFF does not sacrifice the accuracy of many-shot classes.

\subsection{Model Validation}
To further validate the effectiveness of CReFF, we study two key research questions.

\begin{figure}[!t]
	\centering
	\includegraphics[width=.9\linewidth]{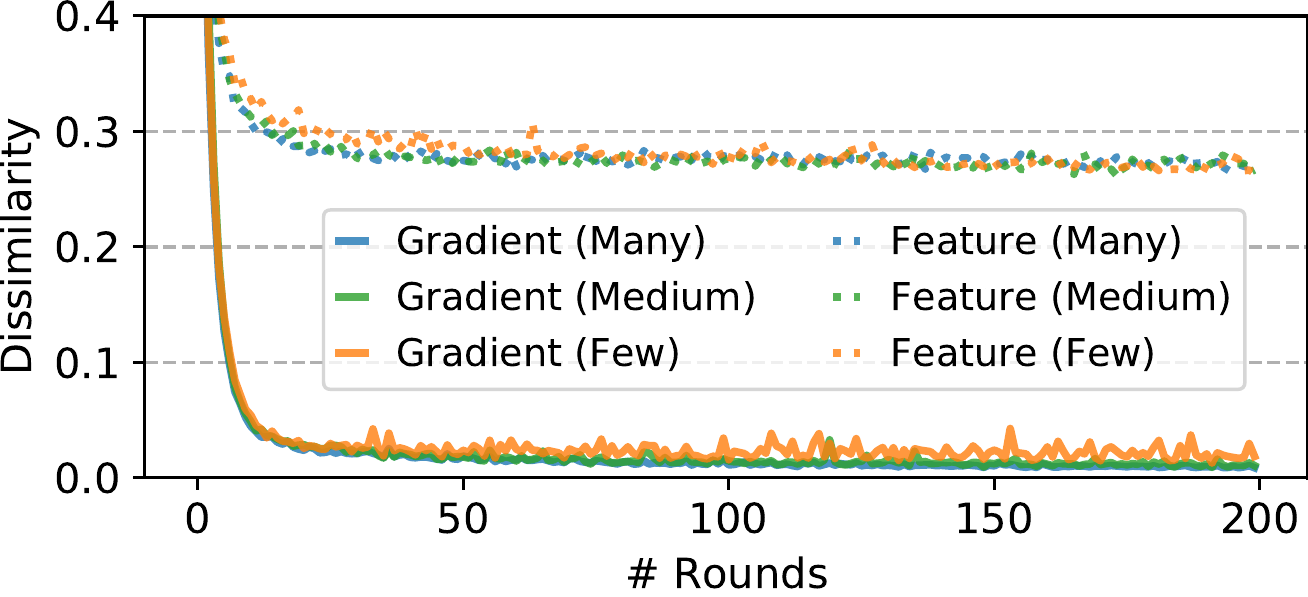}
	\caption{The lines of two dissimilarities on CIFAR-10-LT with $\IF=100$ and $\alpha=0.5$. 
	}
	\label{difference_curve}	
\end{figure}

\noindent\textbf{Why does a classifier re-trained on federated features perform similarly to the one re-trained on real features?}
%%生成特征与原始数据对分类器的梯度方向相同
We show the dissimilarity between the federated and the real feature gradients by the gradient matching loss of each class. We group the classes into three sets for CIFAR-10-LT: Many (more than 1500 samples), medium (200-1500 samples) and few (less than 200 samples), and then average the dissimilarities in each group shown by the solid lines in Figure \ref{difference_curve}. It can be observed that the dissimilarities decrease to around 0 in a few rounds, which means that optimizing on the gradient matching loss successfully makes the federated feature gradients close to the real feature gradients. Thus, the optimization of classifier re-training on federated features follows a similar path as that of the real features. Moreover, we also show the dissimilarity between the federated and the real features by calculating their cosine dissimilarity in Figure \ref{difference_curve} (dotted lines). As training goes on, the dissimilarities between the federated and real features are also decreasing but they are far from 0. It means that the federated features do not have to be similar to the real features although their corresponding gradients are very close. Therefore, the privacy of the real features can be further guaranteed. 

\begin{figure}[!t]
	\centering
		\includegraphics[width=.88\linewidth]{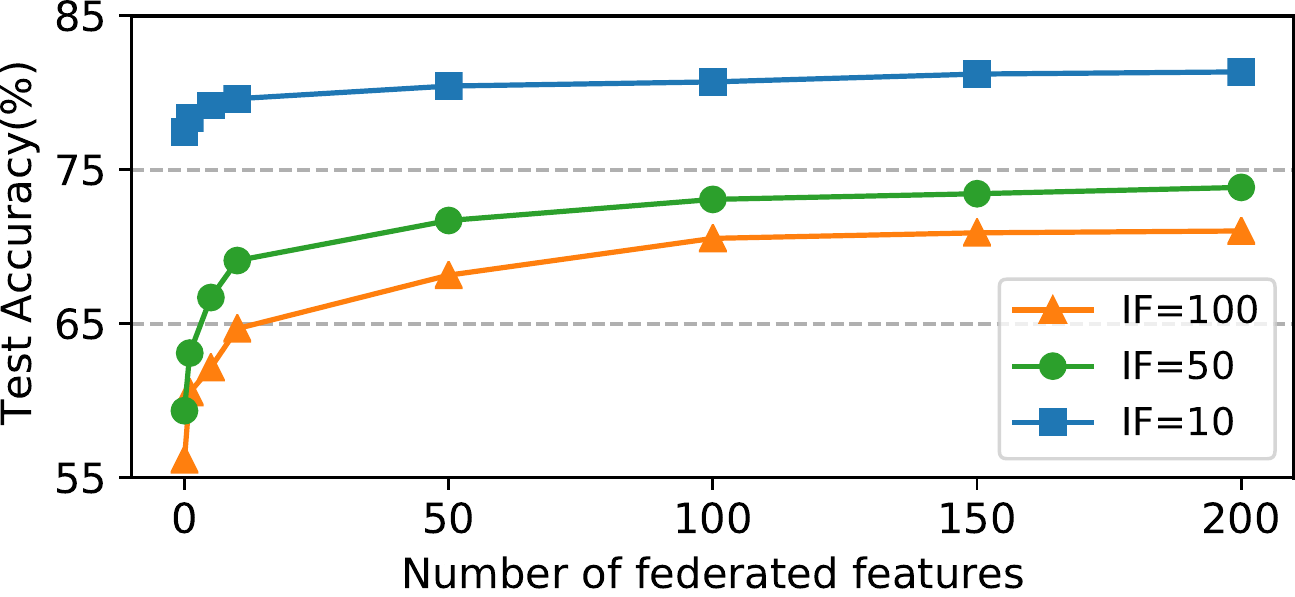}
		\caption{The performance of CReFF with the different number of federated features on CIFAR-10-LT with different $\IF$s.
		}
		\label{num_of_features}	
\end{figure}

\noindent\textbf{How many federated features per class is enough to re-train a good classifier?} One important hyperparameter in CReFF is the number of federated features per class. We evaluate its influence by tuning it in the set $\{0, 1, 5, 10, 50, 100, 150, 200\}$ on CIFAR-10-LT with different $\IF$s. When the number of federated features is 0, CReFF is equivalent to FedAvg. 
It can be observed from Figure \ref{num_of_features} that more federated features produce higher accuracy in general. However, as the number of federated features increases, the improvement gradually becomes stable. This result is anticipated because the information provided by real feature gradients is limited. Meanwhile, a surprising result is that even if only one federated feature is learned for each class, CReFF can still improve 4.21\%, 3.74\%, and 0.92\% for CIFAR-10-LT with $\IF=100$, 50 and 10, respectively.

\section{Conclusion}
In this paper, we first show that classifier re-training is a straightforward and effective way to deal with the joint problem of data heterogeneity and long-tail distribution in the federated learning framework. Based on this, CReFF is proposed to learn a small set of balanced features called \textit{federated features} on the server for classifier re-training in a privacy-preserving manner. The classifier re-trained on the federated features can produce comparable performance as the one retrained on the real data.
%, based on the knowledge from clients in a privacy-preserving manner. 
Experiments have shown that CReFF outperforms the state-of-the-art FL methods in the setting of heterogeneity and long-tail distribution, and the effectiveness CReFF is validated empirically. 
%% The file named.bst is a bibliography style file for BibTeX 0.99c
%\bibliographystyle{named}

\section{Acknowledgments}
We thank Dr. Fei Chen for his helpful discussion and reviewers of IJCAI-ECAI 2022 for their constructive and helpful feedback. This work was supported in part by the National Natural Science Foundation of China (No. 62002302, U21A20514, 61872307 and 52007173), the Open Research Projects of Zhejiang Lab (No. 2021KB0AB03), the Natural Science Foundation of Fujian Province (No. 2020J01005), and Zhejiang Provincial Natural Science Foundation of China (No. LQ20E070002).

\bibliographystyle{named}
\bibliography{fl_ref}

\end{document}